\pdfoutput=1

\documentclass[11pt,a4paper]{article}
\PassOptionsToPackage{hyphens}{url}\usepackage[hyperref]{udst2017}
\usepackage{times}
\usepackage{latexsym}
\usepackage[hyphens]{url}
\usepackage{mathtools}
\usepackage{amsthm}
\usepackage{graphicx}
\usepackage{multirow}
\usepackage{algorithmicx}
\usepackage[noend]{algpseudocode}
\usepackage{algorithm}
\usepackage{esvect}
\usepackage{dsfont}
\usepackage{amsfonts}

\renewcommand{\vec}[1]{\mathbf{#1}}

\algnewcommand{\LeftComment}[1]{\Statex \(\triangleright\) #1}

\aclfinalcopy 

\title{A non-projective greedy dependency parser with bidirectional LSTMs}

\author{David Vilares \\
  Universidade da Coru\~{n}a \\
  LyS Group  \\
  Departamento de Computaci\'{o}n \\
  Campus de Elvi\~{n}a s/n, 15071 \\ A Coru\~{n}a, Spain \\
  {\tt david.vilares@udc.es} \\\And
  Carlos G\'{o}mez-Rodr\'{\i}guez \\
  Universidade da Coru\~{n}a \\
  FASTPARSE Lab, LyS Group \\
  Departamento de Computaci\'{o}n \\
  Campus de A Elvi\~{n}a s/n, 15071 \\ A Coru\~{n}a, Spain \\
  {\tt carlos.gomez@udc.es} \\}

\date{}

\begin{document}
\maketitle
\newcommand{\udst}[0]{\emph{CoNLL 2017 UD Shared Task}}
\begin{abstract}

The LyS-FASTPARSE team presents \textsc{bist-covington}, a neural implementation of the \newcite{covington2001fundamental} algorithm for non-projective dependency parsing. The  bidirectional \textsc{lstm} approach by \newcite{kiperwasser2016simple} is used to train a greedy parser with a dynamic oracle to mitigate error propagation. The model participated in the \udst. In spite of not using any ensemble methods and using the baseline segmentation and PoS tagging, the parser obtained good results on both macro-average LAS and UAS in the \emph{big treebanks} category (55 languages), ranking 7th out of 33 teams. In the \emph{all treebanks} category (LAS and UAS) we ranked 16th and 12th. The gap between the \emph{all} and \emph{big} categories is mainly due to the poor performance on four parallel PUD treebanks, suggesting that some `suffixed' treebanks  (e.g. Spanish-AnCora)  perform poorly on cross-treebank settings, which does not occur with the corresponding `unsuffixed' treebank  (e.g. Spanish). By changing that, we obtain the 11th best LAS among all runs (official and unofficial). The code is made available at \url{https://github.com/CoNLL-UD-2017/LyS-FASTPARSE} 
\end{abstract} 

\section{Introduction}

Dependency parsing is one of the core structured prediction tasks researched by computational linguists, due to the potential advantages that obtaining the syntactic structure of a text has in many natural language processing applications, such as machine translation \cite{miceli2015non,Xiao2016translation}, sentiment analysis \cite{socher2013recursive,vilares2017universal} or information extraction \cite{yu2015combining}.

The goal of a dependency parser is to analyze the syntactic structure of sentences in one or several human languages by obtaining their analyses in the form of dependency trees. Let $w = [w_1,w_2,...,w_{|w|}]$ be an input sentence, a \emph{dependency tree} for $w$ is an edge-labeled directed tree $T=(V,E)$ where $V = \{0,1,2,\ldots,|w|\}$ is the set of nodes and $E = V \times D \times V$ is the set of labeled arcs. Each arc, of the form $(i,d,j)$, corresponds to a syntactic \emph{dependency} between the words $w_i$ and $w_j$; where $i$ is the index of the \emph{head} word, $j$ is the index of the \emph{child} word and $d$ is the \emph{dependency type} representing the kind of syntactic relation between them.\footnote{Following common practice, we are using node $0$ as a dummy root node that acts as the head of the syntactic root(s) of the sentence.} We will write $i \xrightarrow{d} j$ as shorthand for $(i,d,j) \in E$ and we will omit the dependency types when they are not relevant.

\begin{figure}[hbtp]
\begin{center}
   \includegraphics[width=0.79\columnwidth]{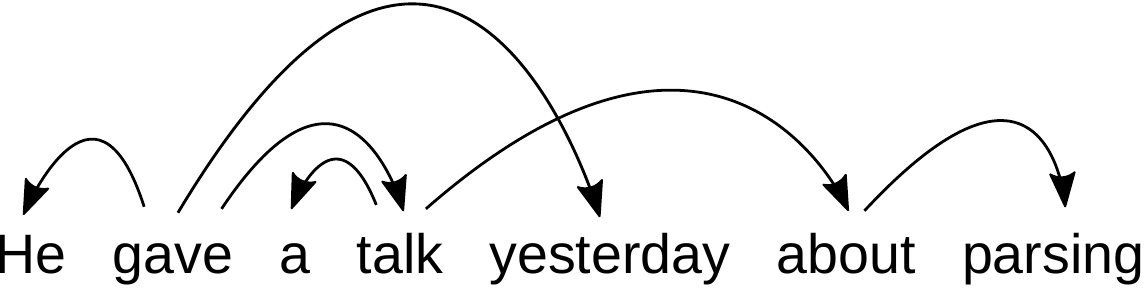}
\caption{\label{figure-non-projective-parsing} A non-projective dependency tree}
\end{center}
\end{figure}

A dependency tree is said to be non-projective if it contains two arcs $i \xrightarrow{} j$ and $k \xrightarrow{} l$ where $min(i,j) < min(k,l) < max(i,j) < max(k,l)$, i.e., if there is any pair of arcs that cross when they are drawn over the sentence, as shown in Figure \ref{figure-non-projective-parsing}. Unrestricted non-projective parsing allows more accurate syntactic representations than projective parsing, but it comes at a higher computational cost, as there is more flexibility in how the tree can be arranged so that more operations are usually needed to explore the much larger search space.

Non-projective transition-based parsing has been actively explored in the last decade \cite{nivre2005pseudo,attardi2006experiments,nivre2008algorithms,nivre2009non,gomez2010transition,gomez2014polynomial}. The success of neural networks and word embeddings for projective dependency parsing \cite{CheMan2014} also encouraged research on neural non-projective models \cite{straka2016ud}. However, to the best of our knowledge, 
no neural implementation is available of unrestricted non-projective transition-based parsing with a dynamic oracle. Here, we present such an implementation for the \newcite{covington2001fundamental} algorithm using bidirectional long short-term memory networks (\textsc{lstm}) \cite{lstm}, which is the main contribution of this paper.

The system is evaluated at the \udst: \emph{end-to-end multilingual parsing using Universal Dependencies} \cite{udst:overview}. The goal is to obtain a Universal Dependencies v2.0 representation \cite{ud} of a collection of raw texts in different languages.

\section{End-to-end multilingual parsing}

Given a raw text, we: (1) segment and tokenize sentences and words, (2) apply part-of-speech (PoS) tagging over them and (3) obtain the dependency structure for each sentence.

\subsection{Segmentation and PoS tagging}

For these two steps we relied on the output provided by UDpipe v1.1 \cite{straka2016ud}, which was provided as a baseline model for the shared task.

\subsection{The \textsc{bist-covington} parser}

\textsc{bist-covington} is built on the top of three core ideas: a non-projective transition-based parsing algorithm \cite{covington2001fundamental,nivre2008algorithms}, a neural scoring model with bidirectional long short-term memory networks as feature extractors that feed a multilayer perceptron \cite{kiperwasser2016simple}, and a dynamic oracle to mitigate error propagation \cite{gomez2efficient}.

\subsubsection{The \newcite{covington2001fundamental} algorithm}\label{section-covington-algorithm}

The idea of Covington's algorithm is quite intuitive: any pair of words $w_{i}$, $w_{j}$ in $w$ have a chance to be connected, so we need to consider all such pairs to determine the type of relation that exists between them (i.e. $i \xrightarrow{d} j$, $j \xrightarrow{d} i$ or \texttt{none}). One pair $(i,j)$ is compared at a time. We will be referring to the indexes $i$ and $j$ as the \emph{focus words}. It is straightforward to conclude that the theoretical complexity of the algorithm is $\mathcal{O}(|w|^2)$. 

Covington's algorithm can be easily implemented as a transition system \cite{nivre2008algorithms}. The set of transitions used in \textsc{bist-covington} and their preconditions is specified in Table \ref{covington-transitions}. 
Each transition corresponds to a parsing configuration represented as a 4-tuple $c$ = $(\lambda_1,\lambda_2, \beta, A)$, such that:
\begin{itemize}
\item $\lambda_1$, $\lambda_2$ are two lists storing the words that have been already processed in previous steps. $\lambda_1$ contains the already processed words for which the parser still has not decided, in the current state, the type of relation with respect to the focus word $j$, located at the top of $\beta$. $\lambda_2$ contains the already processed words for which the parser has already determined the type of relation with respect to $j$ in the current step.
\item $\beta$ contains the words to be processed.
\item $A$ contains the set of arcs already created.
\end{itemize}

\begin{figure}[hbtp]
\begin{center}
   \includegraphics[width=1.0\columnwidth]{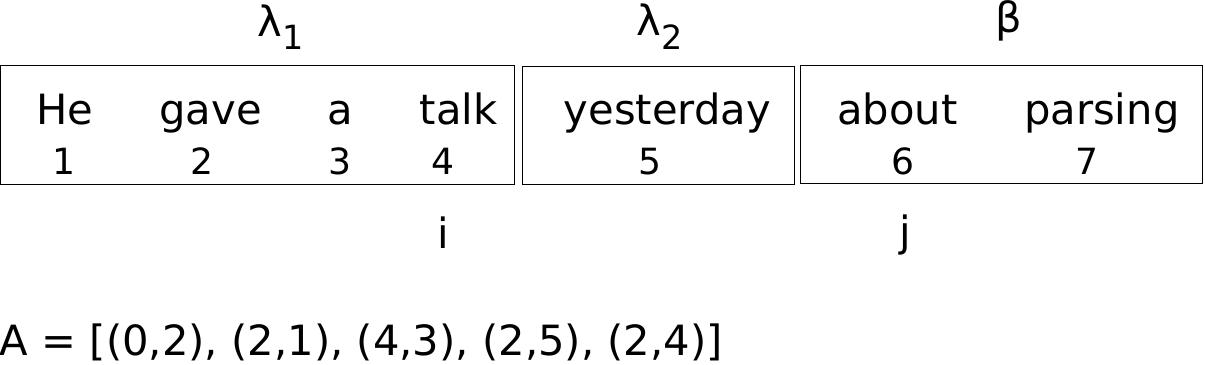}
\caption{\label{figure-middle-parsing-configuration} A parsing configuration for our introductory example just before creating a non-projective \textsc{right arc} $talk \xrightarrow{} about$.}
\end{center}
\end{figure}

Given a sentence $w$ the parser starts at an initial configuration $c_s$ = $([0],[],[1,...,|w|], \{\})$ and will apply valid transitions until reaching a final configuration $c_f$ such that $c_f$ = $(\lambda_1, \lambda_2, [], A)$. 
Figure \ref{figure-middle-parsing-configuration} illustrates an intermediate parsing configuration for our introductory example.

\begin{table*}[h]
\begin{center}
\begin{tabular}{lll}
 \bf Transitions & \bf  & \bf \\ 

{\sc left arc} & $(\lambda_1|i, \lambda_2, j|\beta, A)$ & $(\lambda_1, i| \lambda_2, j|\beta, A \cup \{(j,d,i)\})$  \\

{\sc right arc} & $(\lambda_1|i, \lambda_2,j|\beta,A)$ & $(\lambda_1,i |\lambda_2,j|\beta,A \cup \{(i,d,j)\})$  \\

{\sc shift} & $(\lambda_1,\lambda_2,i|\beta,A)$ & $(\lambda_1 \cdot \lambda_2|i,[],\beta,A)$  \\

{\sc no-arc} & $(\lambda_1|i, \lambda_2,\beta,A)$ & $(\lambda_1 \cdot i|\lambda_2, \beta, A)$ \\
&&\\
\bf Preconditions & & \\
{\sc left arc}  & $i > 0\ and\ \not\exists (k \xrightarrow{} i) \in A\ and\ \not\exists (i \xrightarrow{} ... \xrightarrow{} j)$&\\
{\sc right arc}  & $ \not\exists (k \xrightarrow{} j) \in A\ and\ \not\exists (j \xrightarrow{} ... \xrightarrow{} i) $&\\
{\sc no-arc} &  $i > 0$ &\\

\end{tabular}
\end{center}
\caption{\label{covington-transitions} Set of transitions for \textsc{bist-covington} as described in \newcite{nivre2008algorithms}. $a \xrightarrow{} ... \xrightarrow{} b$ indicates there is a path in the dependency tree that allows to reach $b$ from $a$}
\end{table*}

\subsubsection{A dynamic oracle for Covington's algorithm \cite{gomez2efficient}}\label{section-dynamic-oracle}

Given a gold dependency tree, $\tau_g$, and a parser configuration $c$, we can define a loss function $\mathcal{L}(c,\tau_g)$ that determines the minimum number of missed arcs of $\tau_g$ across the possible outputs ($A$) of final configurations that can be reached from $c$, i.e., the least possible number of errors with respect to $\tau_g$ that we can obtain from $c$. A static (traditional) oracle is only defined on canonical transition sequences that lead to the gold tree, so that $\mathcal{L}(c,\tau_g) = 0$ at every step during the training phase. However, during the test phase such training strategy might end up in serious error propagation, as it is difficult for the parser to recover from wrong configurations that it has never seen, resulting from suboptimal transitions that increase loss. A dynamic oracle \cite{goldberg2012dynamic} explores such wrong configurations during the training phase to overcome this issue. Instead of always picking the optimal transition during training, the parser moves with probability $x$ to an erroneous (loss-increasing) configuration, namely the one with the highest score among those that increase loss.

To compute $\mathcal{L}$ for non-projective trees we used the approach proposed by \newcite[Algorithm 1]{gomez2efficient}. This dynamic oracle can be computed in $\mathcal{O}(|w|)$ although the current implementation in \textsc{bist-covington} is $\mathcal{O}(|w|^3)$. To choose the dependency type corresponding to the selected transition (in case it is a \textsc{left} or {\textsc{right arc}), we look at the gold treebank.

\subsubsection{The \textsc{bist}-parsers  \cite{kiperwasser2016simple}}
\label{sect:thebist}

The original set of \textsc{bist}-parsers is composed of a projective transition-based model using the arc-hybrid algorithm \cite{kuhlmann2011dynamic} and a graph-based model inspired in \newcite{eisner1996three}. They both rely on bidirectional \textsc{lstm}'s (\textsc{bilstm}'s). We kept the main architecture of the arc-hybrid \textsc{bist}-parser and changed the parsing algorithm to that described in \S \ref{section-covington-algorithm} and \S \ref{section-dynamic-oracle}. 
We encourage the reader to consult \newcite{kiperwasser2016simple} for a detailed explanation of their architecture, but we now try to give a quick overview of its use as the core part of \textsc{bist-covington}.\footnote{Including some additional capabilities that we included especially for \textsc{bist-covington}.}

In contrast to traditional parsers \cite{nivre2006maltparser,martins2010turbo,rasooli2015yara}, \textsc{bist}-parsers rely on embeddings as inputs instead of on discrete events (co-occurrences of words, tags, features, etc.). Embeddings are low-dimensional vectors that provide a continuous representation of a linguistic unit (word, PoS tag, etc.) based on its context \cite{mikolov2013distributed}.  

Let $\vec{w}$=$[\vec{w}_1,...,\vec{w}_{|w|}]$ be a list of word embeddings for a sentence, let $\vec{u}$=$[\vec{u}_1,...,\vec{u}_{|w|}]$ be the corresponding list of universal PoS tag embeddings, $\vec{t}$=$[\vec{t}_1,...,\vec{t}_{|w|}]$ the list of specific PoS tag embeddings, $\vec{f}$=$[\vec{f}_1,...,\vec{f}_{|w|}]$  the list of morphological features (``feats'' column in the Universal Dependencies data format) and  $\vec{e}$=$[\vec{e}_1,..., \vec{e}_{|w|}]$ a list of external word embeddings; an input $\vec{x}_i$ for a word $w_i$ to {\textsc{bist-covington}} is defined as:\footnote{ It might turn out that for some treebank/language some of this information is not available, in which case the unavailable elements are considered as empty lists.}

\begin{center}
$\vec{x}_i =  \vec{w}_{i} \circ \vec{u}_{i} \circ \vec{t}_{i}  \circ \vec{f}_{i} \circ \vec{e}_{i}$  
\end{center}

where $\circ$ is the concatenation operator.

Let \textsc{lstm}$(\vec{x})$ be an abstraction of a standard long short-term memory network that processes the sequence $\vec{x}=[\vec{x}_1,...,\vec{x}_{|\vec{x}|}]$, then a \textsc{bilstm} encoding of its $i$th element, \textsc{bilstm}$(\vec{x},i)$ can be defined as:

\begin{center}
\textsc{bilstm}$(\vec{x},i)$ = \textsc{lstm}$(\vec{x}_{1:i})$ $\circ$ \textsc{lstm}$(\vec{x}_{|\vec{x}|:i})$ 
\end{center}

In the case of multilayer \textsc{bilstm's} (\textsc{bist}-parsers allow it), given $n$ layers, the output of the \textsc{bilstm}$_m$ is fed as input to \textsc{bilstm}$_{m+1}$. From the \textsc{bilstm} network we take a hidden vector $\vec{h}$, which can contain the output hidden vectors for: the $x$ leftmost words in $\beta$, the rightmost $y$ of $\lambda_1$, and the $z$ leftmost and $v$ rightmost words in $\lambda_2$.

The hidden vector $\vec{h}$ is used to feed a multilayer perceptron with one hidden layer and four output neurons that predicts which transition to take. The output is computed as $W_2 \cdot tanh(W \cdot \vec{h} + b) + b_2$, where $W, W_2, b$ and $b_2$ correspond to the weight matrices and bias vectors of the hidden and output layer of the perceptron.  Similarly, \textsc{bist}-parsers (including \textsc{bist-covington}) use a second perceptron with one hidden layer to predict the dependency type. In this case the output layer corresponds to the number of dependency types in the training set. 

\subsection{Postprocessing}

\textsc{bist-covington} as it is allows parses with multiple roots, i.e., with several nodes assigned as children of the dummy root. This was not allowed however by the task organizers, as it is enforced by Universal Dependencies that only one word per sentence must depend on the dummy root. To overcome this, the output is postprocessed according to Algorithm \ref{algorithm-single-root}. Basically, we look for the first verb rooted at 0, or for the first word whose head is 0 if there is no verb, and reassign all other words to the selected term:

\begin{algorithm}[h]
\normalsize{
\caption{Multiple to single node root}
\label{algorithm-single-root}
\begin{algorithmic}[1]

\small
\Procedure{to\_single}{V, E}
\LeftComment{Get the nodes rooted at zero (those whose head has to be reassigned)}
\State $RO \gets []$
\For{$i$ in $V$}
\If {$head(i) = 0$}
	\State $append(RO,i)$
\EndIf
\EndFor
\LeftComment{We select the first verb linked to the dummy root to remove multiple roots}

\If {$len(RO) > 1$}
\State $closest\_head \gets RO[0]$
\For {$r0$ in RO}
\If {$utag(r0) =$ \textsc{verb}}
\State $closest\_head \gets r0$
\State break
\EndIf
\EndFor

\LeftComment{Reassign the head of the invalid nodes (rooted to the dummy root) to $closest\_head$}
\For {$r0$ in RO}
\If {$r0 \neq closest\_head$}
\State $head(r0) \gets closest\_head$
\EndIf
\EndFor
\EndIf

\EndProcedure

\end{algorithmic}}
\end{algorithm}

\section{Experiments}

We here describe the official treebanks used in the shared task (\S \ref{section-treebanks}), the general setup used to train the models (\S \ref{section-general-set-up}) and some exceptions to said general setup that were applied to special cases (\S \ref{section-special-configurations}). We also discuss the experimental results obtained by our system in the shared task (\S \ref{section-results}).

\subsection{CoNLL 2017 treebanks}\label{section-treebanks}

\subsubsection{Training/development splits}

60 treebanks from 45 languages were released to train the models, based on Universal Dependencies 2.0 \cite{ud20data}. Most of them already contained official training and development splits. A few others lacked a development set. For these, we applied a training/dev random split (80/20) over the original training set. All development sets were only used to evaluate and tune the trained models. \emph{No development set} was used to train any of the runs, as specified in the task guidelines.

Additionally, four \emph{surprise languages} (truly low resource languages), were considered by the organization for evaluation: Buryat, Kurmanji, North Sami and Upper Sorbian. For these, the organizers only released a tiny sample set consisting of very few sentences annotated according to the \textsc{ud} guidelines. 


\subsubsection{Test splits}

The organizers provided a test split for each of the treebanks released in the training phase, including the surprise languages. Additionally, they provided test sets corresponding to 14 parallel treebanks in different languages translated from a unique source. All of these test sets \cite{ud20testdata} were hidden from the participating teams until the shared task had ended. Using the TIRA environment \cite{tira2} provided for the shared task, participants could execute runs on them, but not see the outputs or the results.

\subsection{General setup}\label{section-general-set-up}

We used the gold training treebanks to train the parsing models. We trained one model per treebank. No predicted training treebank (predicted universal and/or specific tags and morphological features) was  used for training, except for the case of Portuguese (see \S \ref{section-exception-portuguese-model}).\\

\noindent \textbf{Embeddings:} Word embeddings are set to size 100 and universal tag embeddings to 25. Language-specific tag and morphological feature embeddings are used and set to size 25, if they are available for the treebank at hand. Using external word embeddings seems to be beneficial to improve parsing performance \cite{kiperwasser2016simple}, but it also makes models take more time and especially much more memory to train. The external word embeddings used in this work (the ones pretrained by the \udst    \ organizers\footnote{\url{http://hdl.handle.net/11234/1-1989}}) are of size 100. Due to lack of enough computational resources, we only had time to train 38 models (mainly corresponding to the smallest treebanks) including this information. Models trained with external word embeddings are marked in Table \ref{bist-covington-dev-set-results} with $\star$.\\

\noindent\textbf{Parameters}: Adam is used as  optimizer \cite{kingma2014adam}. Models were trained for up to 30 epochs, except for the two smallest training sets (Kazakh and Uyghur), where models were trained for up to 100 epochs. The size of the output of the stacked {\textsc{bilstm}} was set to 512. For very large treebanks (e.g. Czech or Russian-SyntagRus) or treebanks where sentences are very long (e.g. Arabic), we set it to 256, also to counteract the lack of physical resources to finish the task on time. These models are marked in Table \ref{bist-covington-dev-set-results} with $\bullet$. The number of \textsc{bilstm} layers is set to 2. 
To choose a transition, \textsc{bist-covington} looks at the embeddings of: the first word in $\beta$, the rightmost three words in $\lambda_1$, and the leftmost and rightmost word in $\lambda_2$ (i.e., following the notation in Section \ref{sect:thebist}, we set $x=1$, $y=3$, $z=1$ and $v=1$).\\

\noindent \textbf{Other relevant features of the setup}: Aggressive exploration is applied to the dynamic oracle, as in the original arc-hybrid \textsc{bist-parser}.

\subsection{Special cases}\label{section-special-configurations}

For some treebanks, we followed a different strategy due to various issues. We enumerate the changes below:

\subsubsection{The Portuguese model}\label{section-exception-portuguese-model}

Surprisingly, the model trained on the Portuguese treebank suffered a significant loss with respect to the UDpipe baseline when parsing the full predicted (segmentation and tagging) development file. We first hypothesized this was due to a low accuracy on predicting the ``feats'' column in comparison to other languages, as they are pretty sparse. To try to overcome this, we trained a model without considering them, but it did not solve the problem.
Our second option was to train a Portuguese model on its predicted training treebank.\footnote{We used the predicted tokenization and tagging provided by UDpipe.} Additionally, despite being a relative large treebank, we included external word embeddings to boost performance. This helped us to obtain a performance similar to that reported by UDpipe.

\subsubsection{Surprise languages}\label{section-surprise-data}
As training an accurate parser with so little data might be a hard task , especially in the case of \emph{data-hungry} deep learning models, we used other training treebanks for this purpose. We built a set of parsers inspired on the approach presented by \newcite{vilares2016one}, who find that training a multilingual model on merged harmonized treebanks might actually have a positive impact on parsing the corresponding monolingual treebank. In this particular case, we are assuming that a trained model over multilingual treebanks might be able to capture similar treebank structures for unseen languages. 

In particular, we: (1) ran every trained monolingual model on the sample sets, (2) for each surprise language, we chose the top three languages where the corresponding models obtained the best performance and (3) trained a parser taking the first 2\,000 sentences of the training sets corresponding to such languages and merging them.

Thus, we did not use the provided sample data for training, but only as a development set to choose suitable source languages for our cross-lingual approach.

\subsubsection{Parallel (PUD) treebanks }\label{section-parallel-treebansk}

The only information our models knew about the parallel treebanks during the testing phase was the language in which they were written. To parse these languages we follow a simplistic approach, using the models we had already trained on the provided training corpora: (1) if there is only one model trained on the same language we take that model, (2) else if there is more than one model trained on that language, we take the one trained over the largest treebank (in number of sentences), otherwise (3) we parse the PUD treebank using the English model.\footnote{This latter case should and did never happen, as the task organizers specified in advance that the parallel treebanks would correspond to languages with existing treebanks, but we included it as a fallback mechanism.}

\subsection{Results}\label{section-results}

\begin{table}[]
\begin{center}
\small{
\begin{tabular}{|l|l|}
\hline
Treebank&LAS\\
 \hline
 \hline

  Ancient\_Greek &\multirow{2}{*}{67.85$_{8}$}\\ 
  -PROIEL&\\
  Ancient\_Greek&59.83$_{6}$\\ 
  Arabic&66.54$_{10}$\\ 
  Basque&73.27$_{5}$\\ 
  Bulgarian&85.76$_{6}$\\ 
  Catalan&85.37$_{18}$\\ 
  Chinese&56.76$_{2}$\\ 
  Croatian&77.91$_{11}$\\ 
  Czech-CAC&82.71$_{16}$\\ 
  Czech-CLTT&68.92$_{23}$\\ 
  Czech&83.77$_{11}$\\ 
  Danish&75.27$_{11}$\\ 
  Dutch-LassySmall&82.49$_{6}$\\ 
  Dutch           &71.89$_{7}$\\ 
  English-LinES&73.47$_{13}$\\ 
  English-ParTUT&74.50$_{12}$\\ 
  English        &76.00$_{14}$\\ 
  Estonian&61.79$_{7}$\\ 
  Finnish-FTB&76.80$_{7}$\\ 
  Finnish     &76.11$_{8}$\\ 
  French-ParTUT        &72.09$_{25}$\\ 
  French-Sequoia       &77.77$_{23}$\\ 
  French&79.86$_{20}$\\ 
  Galician-TreeGal     &65.42$_{17}$\\ 
  Galician             &79.24$_{12}$\\ 
  German               &68.35$_{22}$\\ 
  Gothic               &62.07$_{7}$\\ 
  Greek                &81.43$_{6}$\\ 
  Hebrew&59.28$_{9}$\\ 
  Hindi                &86.88$_{15}$\\ 
  Hungarian            &66.00$_{9}$\\ 
  Indonesian&72.94$_{23}$\\ 
  Irish                &58.05$_{22}$\\ 
  Italian       &85.60$_{16}$\\ 
  Japanese      &72.68$_{17}$\\ 
  Kazakh               &16.20$_{26}$\\ 
  Korean               &63.85$_{14}$\\ 
  Latin-ITTB           &79.58$_{7}$\\ 
  Latin-PROIEL         &61.45$_{7}$\\ 
  Latin                &48.92$_{7}$\\ 
  Latvian              &63.05$_{7}$\\ 
  Norwegian-Bokmaal    &84.49$_{8}$\\ 
  Norwegian-Nynorsk    &83.10$_{7}$\\ 
  Old\_Church\_Slavonic&67.21$_{4}$\\ 
  Persian              &77.68$_{17}$\\ 
  Polish     &82.09$_{7}$\\ 
  Portuguese-BR   &86.74$_{9}$\\ 
  Portuguese          &80.91$_{19}$\\ 
  Romanian&80.58$_{11}$\\ 
  Russian-SynTagRus&87.55$_{9}$\\ 
  Russian&76.98$_{8}$\\ 
  Slovak&76.47$_{6}$\\ 
  Slovenian-SST        &43.80$_{21}$\\ 
  Slovenian &82.92$_{7}$\\ 
  Spanish-AnCora   &86.83$_{7}$\\ 
  Spanish              &83.24$_{8}$\\ 
  Swedish-LinES&75.04$_{10}$\\ 
  Swedish&77.33$_{13}$\\ 
  Turkish&57.22$_{5}$\\ 
  Ukrainian            &61.21$_{15}$\\ 
  Urdu&78.31$_{9}$\\ 
  Uyghur               &27.92$_{23}$\\ 
  Vietnamese&38.33$_{12}$\\

\hline
\end{tabular}}
\end{center}
\caption{\label{bist-covington-test-set-results} \textsc{bist-covington} results on the test sets, for those treebanks from which a training set was provided (\emph{small} and \emph{big} treebanks categories)}
\end{table}

Official and unofficial results for our model and for the rest of participants on the test set can be found at the task website: \url{http://universaldependencies.org/conll17/results.html}, but in this section we detail the results obtained by \textsc{bist-covington}.

\subsubsection{Results on \emph{small} and \emph{big} treebanks categories}

Table \ref{bist-covington-test-set-results} shows the performance on the test sets for the treebanks where an official training set was released.

In Table \ref{bist-covington-dev-set-results} we summarize our results on the development sets for those treebanks that provided an official one. Although not shown for brevity and clarity reasons, it is easy to check for the reader that \textsc{bist-covington} outperformed the baseline UDpipe\footnote{\url{http://universaldependencies.org/conll17/baseline.html}} for all these treebanks on the gold configuration (gold segmentation, gold tags). The same is true, except for Chinese (-0.69 decrease in LAS) and Portuguese (-0.09), in the fully predicted configuration (end-to-end parsing). It is easy to conclude from the table that including external word embeddings has a positive effect in most of the treebanks we had time to try. This is especially true when performing end-to-end parsing, where only for three languages (English-LinES, Gothic and Old Church Slavonic) a negative effect was observed.\footnote{Due to not so rich embeddings and/or the model finishing earlier than expected during training. See \S \ref{section-issues}.}

\begin{table}[h!]
\tabcolsep=0.1cm
\begin{center}
\scriptsize{
\begin{tabular}{|l|rr|rr|}
\hline
&\multicolumn{2}{c|}{\bf Gold treebank LAS}&\multicolumn{2}{c|}{\bf Predicted treebank LAS}\\
 \bf Treebank & \bf no E& \bf E& \bf no E &\bf E\\
 \hline
 \hline

  Ancient\_Greek &\multirow{2}{*}{81.44}&\multirow{2}{*}{N/A}&\multirow{2}{*}{70.5}&\multirow{2}{*}{N/A}\\ 
  -PROIEL&&&&\\
  Ancient\_Greek$^\star$        &71.01&\bf 71.31&60.41&\bf 61.25\\ 
  Arabic$^{\star\bullet}$               &79.12&\bf 79.71&64.37&\bf 65.62\\ 
  Basque$^\star$               &81.53&\bf 82.06&72.00&\bf 73.42\\ 
  Bulgarian$^\star$            &89.88&\bf 90.46&84.33&\bf 85.30\\ 
  Catalan$^\bullet$            &90.63&N/A&87.21&N/A\\ 
  Chinese              &80.34&N/A&55.31&N/A\\ 
  Croatian$^\star$             &\bf 83.86&83.64&78.04&\bf 78.74\\ 
  Czech-CAC$^\bullet$            & 88.64&N/A&84.93&N/A\\ 
  Czech-CLTT$^\bullet$           &82.28&N/A&68.03&N/A\\ 
  Czech$^\bullet$               &90.70&N/A&85.47&N/A\\ 
  Danish$^\star$               &83.85&\bf 85.78&74.92&\bf 76.94\\ 
  Dutch-LassySmall$^\star$     &86.59&\bf 86.65&76.78&\bf 77.50\\ 
  Dutch                &86.82&N/A&76.47&N/A\\ 
  English-LinES$^\star$        &\bf 83.74&83.05&\bf 76.48&76.44\\ 
  English-ParTUT$^\star$       &84.15&\bf 84.60&76.24&\bf 77.07\\ 
  English              &\bf 88.02&N/A&76.7&N/A\\ 
  Estonian$^\star$             &79.26&\bf 80.21&61.09&\bf 62.80\\ 
  Finnish-FTB          &\bf 89.00&N/A&76.43&N/A\\ 
  Finnish              &86.51&N/A&76.96&N/A\\ 
  French-Sequoia       &89.14&N/A&81.79&N/A\\ 
  French$^\bullet$               &89.86&N/A&85.8&N/A\\ 
  Galician$^{\star\bullet}$             &84.22&82.58&80.17&79.03\\ 
  German               &87.63&N/A&73.61&N/A\\ 
  Gothic$^\star$               &80.82&\bf81.17&\bf60.84&60.82\\ 
  Greek$^\star$                &86.03&\bf86.37&79.74&\bf80.05\\ 
  Hebrew$^{\star\bullet}$               &\bf85.26&85.13&62.18&\bf62.39\\ 
  Hindi                &93.42&N/A&87.41&N/A\\ 
  Hungarian$^\star$            &80.84&\bf81.30&69.16&\bf70.43\\ 
  Indonesian           &80.39&N/A&74.91&N/A\\ 
  Italian-ParTUT$^\star$       &86.20&\bf86.83&78.90&\bf79.56\\ 
  Italian              &90.30&N/A&86.05&N/A\\ 
  Japanese$^\star$            &\bf96.48&96.46&73.99&\bf74.20\\ 
  Korean               &68.66&N/A&60.18&N/A\\ 
  Latin-ITTB           &84.21&N/A&72.22&N/A\\ 
  Latin-PROIEL         &79.37&N/A&61.98&N/A\\ 
  Latvian$^\star$              &\bf77.25&76.55&63.12&\bf63.62\\ 
  Norwegian-Bokmaal    &91.45&N/A&85.13&N/A\\ 
  Norwegian-Nynorsk    &91.06&N/A&83.38&N/A\\ 
  Old\_Church\_Slavonic$^\star$  &\bf84.59&84.52&\bf66.93&66.66\\ 
  Persian$^{\star\bullet}$              &86.85&N/A&80.44&81.45\\ 
  Polish$^\star$               &91.04&\bf91.25&81.43&\bf82.18\\ 
  Portuguese-BR$^\bullet$        &90.91&N/A&86.41&N/A\\ 
  Portuguese$^{\star\bullet}$           &\bf94.94&93.09&79.3&\bf84.00\\ 
  Romanian$^\star$             &\bf 85.08&84.44&80.97&\bf 81.01\\ 
  Russian-SynTagRus$^\bullet$    &91.91&N/A&88.29&N/A\\ 
  Russian$^\star$              &85.12&\bf86.07&78.02&\bf79.09\\ 
  Slovak$^\star$              &87.61&\bf88.39&75.59&\bf77.35\\ 
  Slovenian$^\star$            &92.28&\bf93.14&82.48&\bf84.15\\ 
  Spanish-AnCora$^\bullet$      &90.50&N/A&86.21&N/A\\ 
  Spanish$^\bullet$              &87.90&N/A&84.25&N/A\\ 
  Swedish-LinES$^\star$        &84.23&\bf84.44&76.39&\bf76.86\\ 
  Swedish$^\star$            &84.88&\bf85.03&76.41&\bf76.64\\ 
  Turkish$^\star$             &61.66&\bf64.46&55.05&\bf57.60\\ 
  Urdu$^\star$                 &\bf87.63&87.50&77.43&\bf77.49\\ 
  Vietnamese$^\star$           &72.21&\bf72.58&42.27&\bf42.94\\

\hline
\end{tabular}}
\end{center}
\caption{\label{bist-covington-dev-set-results} \textsc{bist-covington} results on the dev set, for those treebanks that have an official dev set (all treebanks except French-ParTUT, Irish, Galician-TreeGal, Kazakh, Slovenian-SST, Kazakh, Uyghur and Ukrainian). $\star$ indicates the model was also trained with external word embeddings (E). $\bullet$ indicates the \textsc{bilstm} output dimension was 256. The performance of some models is likely to be improved, as its training finished earlier than expected due to lack of time to finish it or memory issues (see also \S \ref{section-issues})}
\end{table}

Table \ref{suprise-languages-sample-performance} shows the top three selected languages for each surprise treebank, the performance of the monolingual and multilingual (merged) models on them on the sample set (used as dev set), and also shows the performance of the multilingual models in the official test sets.

\begin{table}[h]
\begin{center}
\tabcolsep=0.05cm
\small{
\begin{tabular}{|l|lcc|c|}
\hline
 \bf Surprise & \bf Top 3& {\bf Sample set} & {\bf Sample set} & \bf {Test set}\\
 
 \bf language&\bf treebanks&\bf \emph{Monolingual}  & \bf \emph{Multilingual} &\bf \emph{Multi} \\
\hline
\hline
 \multirow{3}{*}{Buryat} &Hindi&36.60&\multirow{3}{*}{ 43.14}&\multirow{3}{*}{28.65$_{5}$}\\
        & German&32.68&&\\
        & Korean&27.45&&\\
 \hline
 \multirow{3}{*}{Kurmanji}& Romanian&38.84&\multirow{3}{*}{ 39.26}&\multirow{3}{*}{32.08$_{16}$}\\
         & Czech&37.19&&\\
         & Slovenian&31.40&&\\
  \hline
 \multirow{2}{*}{North} &Estonian&45.38&\multirow{3}{*}{ 57.14}&\multirow{3}{*}{32.58$_{14}$}\\
           &Finnish&40.82&&\\
           Sami&Finnish-FTB&40.14&&\\
  \hline
 \multirow{2}{*}{Upper}&Slovenian&65.22&\multirow{3}{*}{ 70.65}&\multirow{3}{*}{52.50$_{15}$}\\
              &Slovak&64.78&&\\
              Sorbian&Bulgarian&61.09&&\\
\hline
\end{tabular}}
\end{center}
\caption{\label{suprise-languages-sample-performance} LAS on the surprise languages sample sets for: (1) top 3 best performing monolingual models for which there is an official training treebank and (2) a multilingual model trained on the first 2\,000 sentences of each of such treebanks. For the multilingual models, the last column shows its performance on the test sets (subscripts indicate our ranking in that language)}
\end{table}

Table \ref{pud-languages-sample-performance}} shows our performance on the PUD treebanks (test sets). There are 4 PUD treebanks for which we obtained a poor performance: Spanish, Finnish, Portuguese and Russian. Average LAS loss with respect to the top system in the corresponding treebank was 32.47, which implied a LAS loss up to 1.60 points in the official global ranking. We hypothesized that taking the model trained on the largest treebank of the same language was the safest option to parse PUD texts, but in retrospective this clearly was not the optimal choice. Those four PUD treebanks were parsed with models trained on Universal Dependencies (UD) treebanks whose official name has a \emph{suffix} (i.e. Spanish-Ancora, Finnish-FTB, Portuguese-BR and Russian-SyntagRus), which were larger than the unsuffixed UD treebank. However, we think such a poor performance surpasses what can be reasonably expected from an universal treebank written in the same language. 
From Table \ref{pud-languages-sample-performance} it is reasonable to conclude that such suffixed treebanks parse more than poorly on cross-treebank settings, in comparison to the model trained on the unsuffixed treebank (rightmost column). We wonder if this can be an indicator of those treebanks sharing universal dependency types, but diverging in terms of syntactic structures, which caused the low LAS scores in those cases. 

A possible contributing factor to this could be that the annotators of the parallel treebanks used guidelines from the unsuffixed treebanks, or automatic output trained on them, as a starting point from the annotation process. At the point of writing we cannot confirm whether this is the case, as documentation for the PUD treebanks is not yet publicly available.

\begin{table}[h]
\begin{center}
\tabcolsep=0.13cm
\scriptsize{
\begin{tabular}{|l|ll||ll|}
\hline
\bf PUD &\bf Trained on &\bf  \multirow{3}{*}{\bf LAS}&\bf Trained on&  \multirow{3}{*}{\bf LAS}\\
\bf treebank&\bf largest treebank&&\bf uns. treebank&\\
&\bf (official)&&\bf(unofficial)&\\
\hline
\hline

Arabic&Arabic&45.12$_{11}$& \multicolumn{2}{c|}{=}\\
Czech&Czech&80.13$_{10}$&\multicolumn{2}{c|}{=}\\
German&German&66.29$_{19}$&\multicolumn{2}{c|}{=}\\
English&English&78.79$_{16}$&\multicolumn{2}{c|}{=}\\
Spanish&Spanish-Ancora&53.73$_{30}$$\downarrow$&Spanish&78.90\\
Finnish&Finnish-FTB&40.66$_{28}$ $\downarrow$&Finnish&80.70\\
French&French&73.15$_{23}$&\multicolumn{2}{c|}{=}\\
Hindi&Hindi&51.15$_{13}$&\multicolumn{2}{c|}{=}\\
Italian&Italian&83.84$_{15}$&\multicolumn{2}{c|}{=}\\
Japanese&Japanese&76.09$_{18}$&\multicolumn{2}{c|}{=}\\
Portuguese&Portuguese-BR&$54.75_{27}$ $\downarrow$&Portuguese&72.84\\
Russian&Russian-SyntagRus&44.69$_{31}$ $\downarrow$&Russian&70.00\\
Swedish&Swedish&69.60$_{17}$&\multicolumn{2}{c|}{=}\\
Turkish&Turkish&34.96$_{4}$&\multicolumn{2}{c|}{=}\\

\hline
\end{tabular}}
\end{center}
\caption{\label{pud-languages-sample-performance} LAS/UAS performance on the PUD treebanks (test sets). The $\downarrow$ symbol indicates a drastic gap in performance with respect the average performance of \textsc{bist-covington}. We show how parsing the PUD treebank with a model trained on the corresponding unsuffixed treebank clearly improves the LAS accuracy.}
\end{table}

\section{Discussion}

\textsc{bist-covington} worked very well on languages where official training/development sets were available, what the organizers named \emph{big treebanks} (55 treebanks), category where we ranked 7th out of 33 systems, both for LAS and UAS metrics, in spite of not using any ensemble method and not performing custom tokenization, segmentation or tagging.

More in detail, we ranked in the top ten LAS for 35 languages, where 32 belong to the category of \emph{big treebanks}: Arabic (10th), Bulgarian (6th), Buryat (5th), Czech-PUD (10th), Old Church Slavonic (4th), Greek (6th), Spanish (8th), Spanish-Ancora (7th), Estonian (7th), Basque (5th), Finnish (8th), Finnish-ftb (7th), Gothic (7th), Ancient\_Greek (6th), Ancient\_Greek-PROIEL (8th), Hebrew (9th), Hungarian (9th), Latin (7th), Latin-ITB (7th), Latin-PROIEL (7th), Latvian (7th), Dutch (7th), Dutch-lassysmall (6th), Norwegian-Bokmaal (8th), Norwegian-Nynorsk (7th), Polish (7th), Portuguese-BR (9th), Russian (8th), Russian-Syntagrus(9th), Slovak (6th), Slovenian (7th), Swedish-LinES (10th), Turkish (5th), Turkish-PUD (4th) and Ukrainian (9th).

We failed on a subset of the PUD treebanks. As previously explained, the main gap came from the Spanish, Russian, Portuguese and Finnish PUD treebanks. We analyzed those treebanks based on existing UD CoNLL treebanks. We parsed them with the model trained on the largest treebank that shared the language. It turned out that those PUD treebanks that were parsed with suffixed treebanks (e.g. Spanish-Ancora or Russian-SynTagRus) obtained a very low performance, something that did not happen when parsing them with the model trained on the corresponding unsuffixed treebank (e.g. Spanish or Russian). In cases where there was only one UD treebank sharing the language, our approach worked reasonably well, in spite of the simplistic strategy followed (e.g. Turkish-PUD or Czech-PUD).

We did not perform too well either on the set of \emph{small treebanks} (French-ParTUT, Irish, Galician-TreeGal, Kazakh, Slovenian-SST, Uyghur and Ukrainian). This was somewhat expected for two reasons: (1) neural models that are fed with continuous vector representations are usually data-hungry and (2) the submitted model was only trained on our training split; we did not include the \emph{ad-hoc} dev sets for those languages as a part of the final training data.  

We believe that the cases where the parser did not work well were due to external causes (e.g. the chosen cross-treebank strategy), as shown in the case of the PUD treebanks. Unofficial results such as the ones in Table \ref{pud-languages-sample-performance} show that this can be easily addressed to push \textsc{bist-covington} to obtain competitive results in those treebanks too. 

\section{Hardware requirements and issues}\label{section-issues}

Our models required \href{https://github.com/clab/dynet}{DyNet} \cite{dynet}, which allocates memory when it is launched. We ran them on CPU. To train the models we used two servers with 128GB of RAM memory each. Estimating the required memory to allocate to train each model was a hard task for us. Dynet does not currently have a garbage collector,\footnote{\url{https://github.com/clab/dynet/issues/418}} so many models ran out of memory even before finishing their training, probably due to wrong memory estimations to complete this phase, and our lack of resources to allocate memory for many treebanks at a time. We observed that models such as Arabic with external word embeddings could take up to 64GB during the training phase. 

The performance on the dev set of our trained models was close, but not equal, in our training machine and in TIRA. This might be caused by a serialization versioning issue: \url{https://github.com/clab/dynet/issues/84}.

To safely run a large trained model with external embeddings we recommend at least 32GB of RAM memory. We think a safe estimate to run any model without external embeddings would be something between 15 and 20GB.

The current version of \textsc{bist-covington} is not very fast. Average speed (tokens/second) over all test treebanks was 18.27. The fastest models were Kazakh (66.36), Uyghur (54.11) and Czech-PUD (45.79) and the slowest ones Czech-CLTT (5.37), Latin-PROIEL (7.69) and Galician-TreeGal (8.19). To complete the testing phase of the shared task, \textsc{bist-covington} took around 28 hours. These times correspond to those of the official evaluation on the TIRA virtual machine. Several factors influence these speeds. Firstly, RNN approaches tend to be slower than feedforward approaches (e.g., reported speeds for the original transition-based \textsc{bist}-parser by \citet{kiperwasser2016simple} are an order of magnitude behind those of \citet{CheMan2014}, although the latter is also much less accurate). Secondly, parsing UD data for different languages accurately requires using more linguistic information (e.g. feature embeddings), increasing the model size with respect to models evaluated on simpler settings like the English Penn Treebank. Finally, we are aware that Covington's algorithm may become slower when sentences are too long due to its quadratic worst-case complexity, an issue that is likely to happen due to the predicted segmentation (the organizers actually informed that some treebanks contained sentences of about 300 words).

\section{Conclusion}

This paper presented \textsc{bist-covington}, a bidirectional \textsc{lstm} implementation of the \newcite{covington2001fundamental} algorithm for non-projective transition-based dependency parsing. Our model was evaluated on the end-to-end multilingual parsing with universal dependencies shared task proposed at CoNLL 2017. For segmentation and part-of-speech tagging our model relied on the official UDpipe baseline. The official results located us 7th out of 33 teams in the \emph{big treebanks} category, in spite of not using any ensemble method.

As future work, there is room for improvement. Due to lack of resources to train the models and complete the task on time, we could not train all models using external word embeddings, which has been shown to produce a significant overall improvement. Jackniffing \cite{agic2017not} might be a simple way to improve the LAS scores. Finally, it would be interesting to implement the non-monotonic version of the Covington transition system, together with approximate dynamic oracles \cite{nonmonotonic}, shown to improve accuracy over the regular Covington parser.

\section*{Acknowledgments}

David Vilares is funded by an
FPU Grant 13/01180. Carlos G{\'o}mez-Rodr{\'i}guez has received funding from the European Research Council (ERC), under the European Union's Horizon 2020 research and innovation programme (FASTPARSE, grant agreement No 714150). Both authors have received funding from the TELEPARES-UDC project  from MINECO. 

\bibliography{acl2017}
\bibliographystyle{acl_natbib}

\end{document}